\pdfoutput=1

\documentclass[11pt]{article}

\usepackage[final]{acl}

\usepackage{times}
\usepackage{latexsym}

\usepackage[T1]{fontenc}

\usepackage[utf8]{inputenc}

\usepackage{microtype}

\usepackage{inconsolata}

\usepackage{eccvabbrv}
\usepackage{graphicx}

\usepackage{amsmath}
\usepackage{multirow}
\usepackage{xspace}
\usepackage{cleveref}
\usepackage{booktabs}
\usepackage{bbding}

\newcommand{\logo}[0]{\text{\smash{\raisebox{-4pt}{\includegraphics[height=14pt]{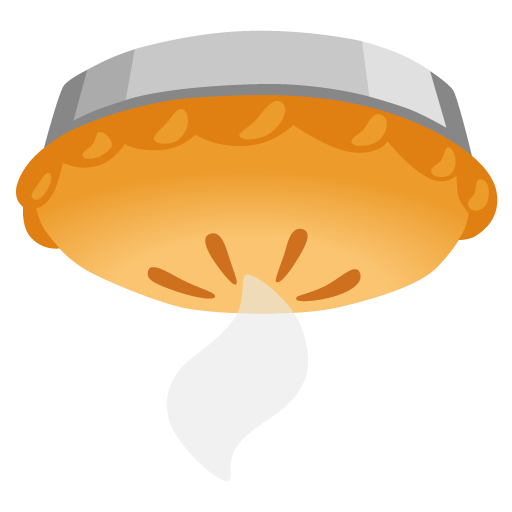}}}}\xspace}

\newcommand{\modelnamelong}{Understanding Pun with Image Explanations\xspace}
\newcommand{\modelname}{UNPIE\xspace}
\newcommand{\modelnamefancy}{\hspace*{2pt}\logo UNPIE\xspace}  

\newcommand{\bfsection}[1]{\noindent\textbf{#1}.}
\newcommand{\bfsectionnodot}[1]{\noindent\textbf{#1}}
\usepackage{listings}
\usepackage{xcolor}
\usepackage{tcolorbox}
\usepackage{fancyvrb}
\definecolor{codegreen}{rgb}{0,0.6,0}
\definecolor{codegray}{rgb}{0.5,0.5,0.5}
\definecolor{codepurple}{rgb}{0.58,0,0.82}
\definecolor{backcolour}{rgb}{0.95,0.95,0.92}

\title{Can visual language models resolve textual ambiguity with visual cues?\\Let visual puns tell you!}

\author{
  \textbf{Jiwan Chung} \quad
  \textbf{Seungwon Lim} \quad
  \textbf{Jaehyun Jeon} \quad
  \textbf{Seungbeen Lee} \quad
  \textbf{Youngjae Yu}
\\
  Yonsei University
\\
  \small{
    \href{mailto:email@domain}{jiwan.chung.research@gmail.com}
  }
}

\begin{document}

\maketitle
\begin{abstract}
Humans possess \textit{multimodal literacy}, allowing them to actively integrate information from various modalities to form reasoning. Faced with challenges like lexical ambiguity in text, we supplement this with other modalities, such as thumbnail images or textbook illustrations. Is it possible for machines to achieve a similar multimodal understanding capability?

In response, we present \textit{\modelnamelong} (\modelnamefancy)\footnote{Data: \url{huggingface.co/datasets/jiwan-chung/VisualPun_UNPIE}}, a novel benchmark designed to assess the impact of multimodal inputs in resolving lexical ambiguities. Puns serve as the ideal subject for this evaluation due to their intrinsic ambiguity.
Our dataset includes 1,000 puns, each accompanied by an image that explains both meanings.
We pose three multimodal challenges with the annotations to assess different aspects of multimodal literacy; Pun Grounding, Disambiguation, and Reconstruction.
The results\footnote{Code: \url{github.com/JiwanChung/VisualPun_UNPIE}} indicate that various Socratic Models and Visual-Language Models improve over the text-only models when given visual context, particularly as the complexity of the tasks increases.
\end{abstract}    

\section{Introduction}
\label{sec:intro}

Humans can actively integrate information from multimodal sources without being explicitly told to.
For example, a wink can reveal the insincerity behind a statement about dieting. Similarly, visual aids such as Venn diagrams help students understand abstract concepts such as set theory.
This active understanding capacity is often denoted as \textit{multimodal literacy}~\cite{mills2017multimodal}.

In contrast, current multimodal models lack this capacity for active understanding and typically operate under two assumptions: (1) all instructions require visual inputs, and (2) these inputs are relevant~\cite{cui2023holistic, zhang2024benchmarking}. Such limitations hinder their applicability in real-world scenarios, such as summarizing long blog posts, where irrelevant images must be excluded, and only contextually significant visuals should be used to enhance the understanding of disparate text segments.

\begin{figure}
    \centering
    \includegraphics[width=0.48\textwidth]{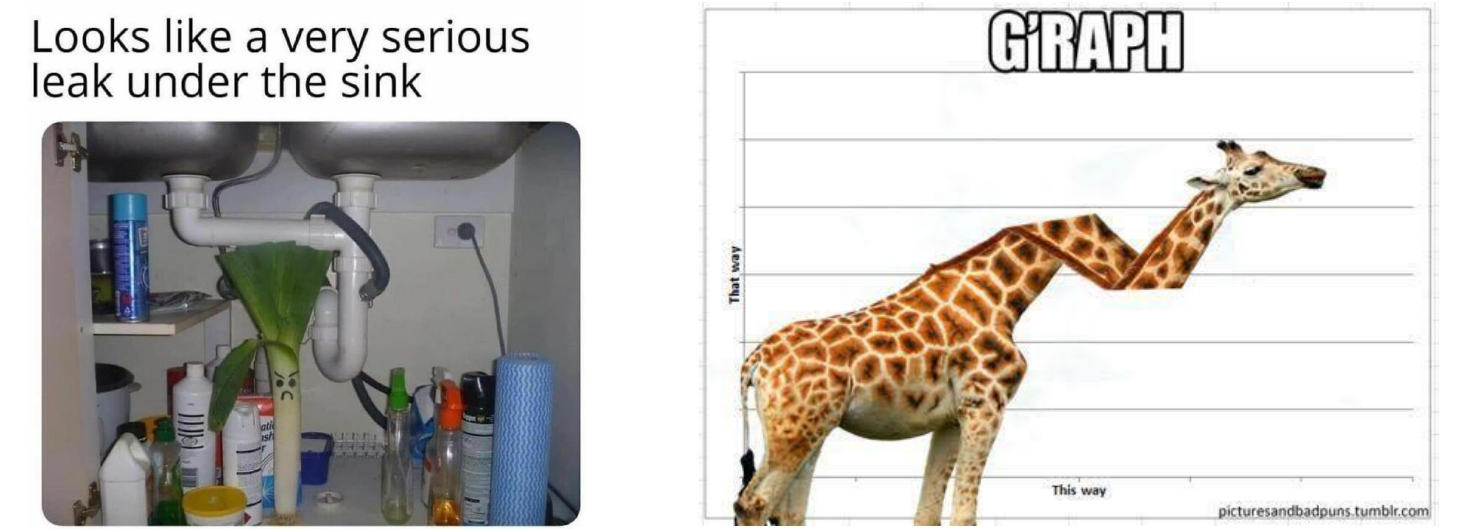}
    \caption{
    Puns naturally occur with images to enhance understanding~\cite{zenner2018one}, making them natural candidates for testing active multimodal understanding capacity of machines.
    Examples of puns accompanied by visual explanations from  \textit{r/puns} subreddit on Reddit.
    }
    \label{fig:intuition}
\end{figure}
An essential component of multimodal literacy is the ability to resolve multimodal ambiguities effectively, which refers to the capacity to disambiguate conflicting or unclear information in modality with information from another modality~\cite{kottur2021simmc,guo2022gravl}.
Owing to its explicit requirement of multimodal information gathering, disambiguation can serve as a controlled benchmark for evaluating multimodal literacy.

\begin{figure*}[ht]
    \centering
    \includegraphics[width=1.0\textwidth]{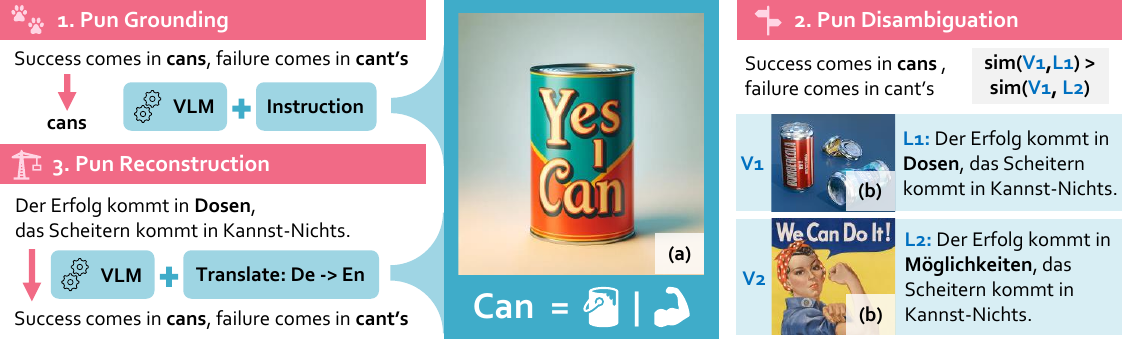}
    \caption{
    The \modelname benchmark comprises three multimodal tasks: 1. Identifying the specific phrase in an English sentence that constitutes a pun, using the provided (a) \textit{pun explanation} image; 2. Choosing the translation of the pun sentence that aligns more closely with the given (b) \textit{pun disambiguator} image; and 3. Reconstructing the English pun sentence from its translated version, aided by the corresponding (a) \textit{pun explanation} image.
    }
    \label{fig:main}
\end{figure*}
Puns stand as a unique challenge within ambiguity modeling. 
They are \textit{intrinsically ambiguous} and understanding a pun requires grasping multiple interpretations of a single phrase or word simultaneously. Understanding puns can be difficult even for humans, often necessitating visual cues to clarify the intended interpretation, as demonstrated in Figure~\ref{fig:intuition}. Compared to verbose textual explanations, visual cues can deliver instant insight, preserving the humor and cleverness of the pun~\cite{morreall1983taking}. Therefore, puns provide an ideal testing ground for assessing models' capabilities in multimodal interpretation.

In this work, we explore model capabilities in \textit{resolving textual ambiguities through visual context}.
To this end, we propose \modelnamelong (\modelnamefancy),
a novel benchmark consisting of 1,000 text-based puns paired with illustrative images that highlight the incongruity within the puns.
Additionally, our dataset approaches pun comprehension as a translation task with incomplete information. This method provides a tangible way to measure the often subjective skill of reconstructing puns.
Each English pun is accompanied by translations in three different languages—German, French, and Korean—to capture the challenge of reconstructing the puns across diverse linguistic contexts.

We design three tests based on \modelname to study how models can exploit visual context to aid pun understanding.
Figure~\ref{fig:main} summarizes the tasks comprising our benchmark.
We first consider an English-only task \textit{pun grounding} that challenges machines to identify the specific phrase in a sentence that forms a pun. Next, we formulate a multilingual challenge of \textit{pun disambiguation} where models must choose the translation that best matches the image provided as a pun disambiguator.
The final test, \textit{pun reconstruction}, is a comprehensive task where models should recreate the original English pun sentence using a translated version with potentially no ambiguity.
For both the pun grounding and reconstruction tasks, we additionally provide the \textit{pun explanation} images as inputs to verify whether models can consider multimodal context when dealing with ambiguous text.

Our comprehensive experiments on \modelname affirm the presence of multimodal literacy capacity in two model types: monolithic Visual-Language Models and modular Socratic Models.
Incorporating visual context consistently improved performance across our three pun comprehension tests. Notably, this improvement was more pronounced in more challenging tasks.
Moreover, VLMs performed better than Socratic Models built on simple image captions.
The result suggests that detailed visual understanding is necessary in our benchmark. 
Finally, fine-tuning with a standard multimodal machine translation dataset adversely affects performance in the pun reconstruction task. 
This degradation aligns with findings from prior studies~\cite{futeral-etal-2023-tackling} stating that web-based multimodal translation datasets may not effectively capture visual dependencies.

Overall, our contributions are as follows:
\begin{enumerate}
    \item \modelnamefancy, a novel benchmark for assessing the multimodal literacy capability of visual-language models. \modelname is built on text with intrinsic ambiguity (puns), guaranteeing the benefit of visual context.
    \item Three new tasks posed on the textual puns and the image annotations: pun grounding, disambiguation, and reconstruction.
    \item Experimental results verifying multimodal literacy capability of both VLMs and Socratic Models concerning pun understanding.
\end{enumerate}
\begin{figure}[ht]
    \centering
    \includegraphics[width=0.48\textwidth]{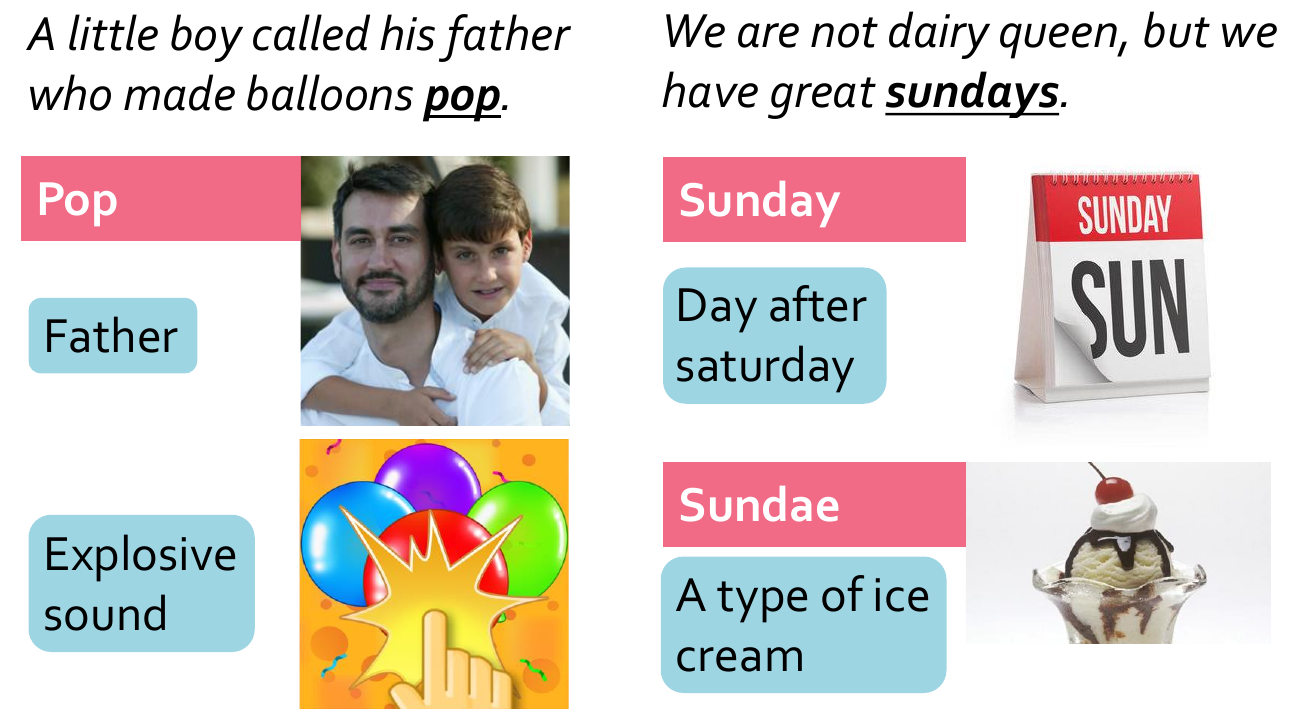}
    \caption{Comparison of homographic (left) and heterographic (right) puns in \modelname dataset along with the respective \textit{disambiguator} visual annotations.}
    \label{fig:homo_hetero}
\end{figure}

\begin{table}[t]
    \centering
    \begin{tabular}{lccc}
    \toprule
    Dataset & Size & Ambiguous (\%) & Gen \\
    \midrule
    Multi30k & 1000 & 2 \%  & \Checkmark \\
    CoMMuTE & 155 & 100 \%  &  \\
    \modelnamefancy & 1000 & 100 \%  & \Checkmark \\
    \bottomrule
    \end{tabular}
    \caption{Comparison of \modelname against multimodal machine translation benchmarks. The statistics for Multi30k are from the \textit{test-2017-flickr} subset. Gen denotes a generative benchmark.}
    \label{tab:data_stats}
\end{table}

\begin{figure}[ht]
    \centering
    \includegraphics[width=0.48\textwidth]{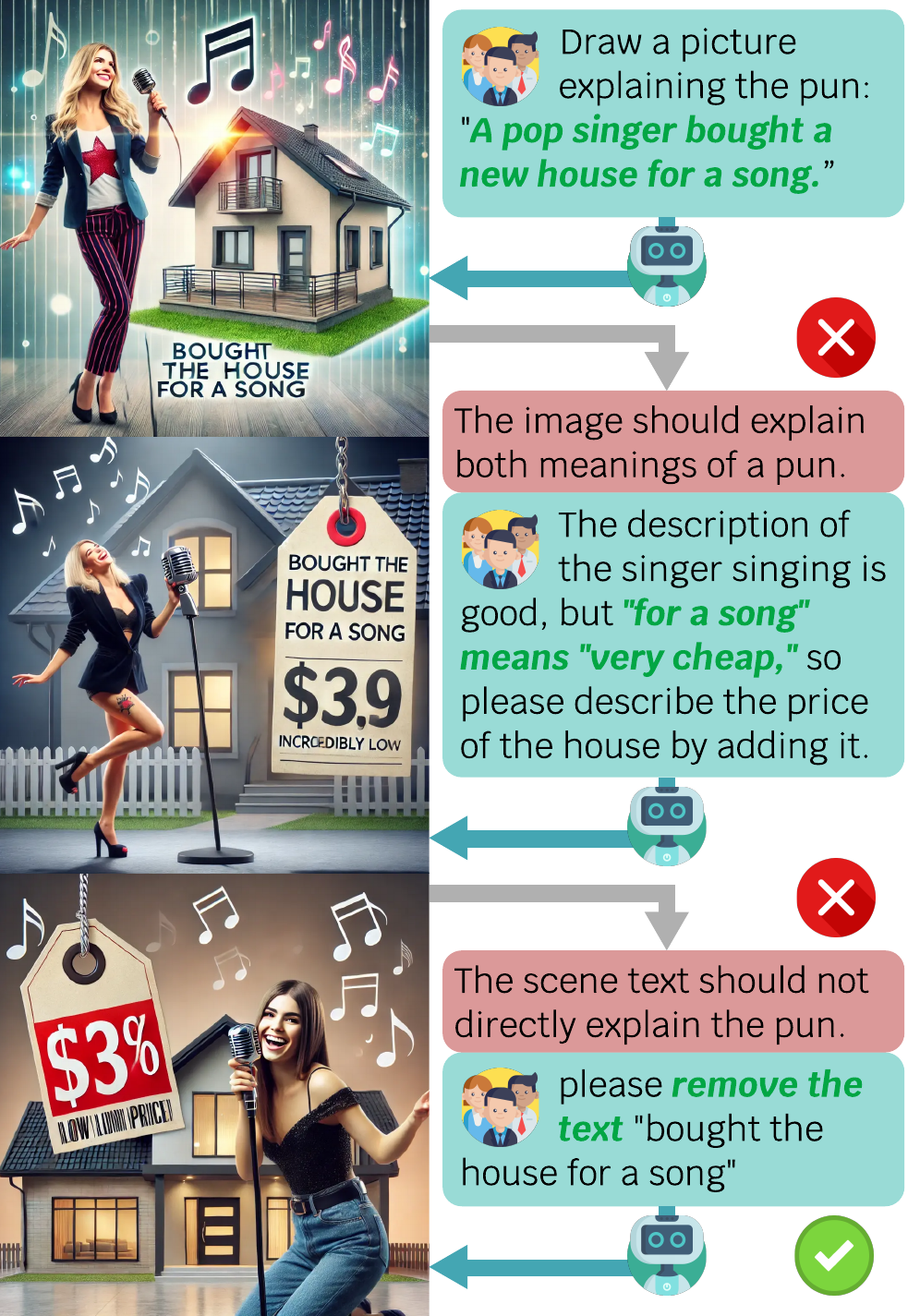}
    \caption{An example of our \textit{pun explanation} image generation process. A human worker interacts with an off-the-shelf text-to-image model, iteratively guiding the model to produce an image that satisfies each specified criterion.}
    \label{fig:image_gen_process}
\end{figure}

\section{Overview of \modelnamefancy Benchmark}
\label{sec:ours_bench}

\modelname is a new multimodal multilingual benchmark.
Its primary aim is to assess machines' capacity to actively integrate information from visual sources to resolve ambiguity in text.
Our dataset leverages puns that inherently contain such ambiguity to study the challenge of multimodal literacy in a natural environment.

\modelname extends puns in two directions: visual context and multilingual translations.
First, we collect images for each pun that 1. describes both meanings of the pun to explain it and 2. depicts only one meaning of the pun to disambiguate the pun  (\cref{subsec:data_collection_vis}).
While one can naturally retrieve images for disambiguation from the web, images that illustrate the ambiguity of the pun in a single canvas are rare.
Thus, we use an off-the-shelf text-to-image model~\cite{openai2023dalle3} to generate such images.
We then employ \textit{human annotators} to filter the images so that they correctly explain the given pun.
Secondly, we ask human annotators to translate the English pun sentences into multilingual targets (\cref{subsec:data_collection_translation}).
Importantly, the ambiguity should not carry on to the translation target.

\subsection{Collecting Puns with Visual Context}
\label{subsec:data_collection_vis}

\bfsection{Base Text-Only Pun Data}
We build our multimodal multilingual benchmark on top of the text-only English pun dataset of SemEval 2017 Task 7~\cite{miller2017semeval}. The dataset bounds the pun understanding problem in two ways to rely less on external requirements: first, each sentence contains a maximum of one pun.
Hence, a sentence's lexical ambiguity is regulated, at least in terms of puns.
Second, most pun has a lexical entry in WordNet 3.1 (81\% of the whole data). This vocabulary limit keeps our pun generation problem from being dominated by many out-of-vocabulary words.

The data is divided into Homographic and Heterographic puns, depending on the surface form of the puns.
As shown in Figure~\ref{fig:homo_hetero}, homographic puns have identical spelling and pronunciation but different meanings, while heterographic puns differ in spelling and meanings.
We inherit this categorization scheme and report our experiment results category-wise (Homographic and Heterographic).
From the SemEval 2017 collection of 2,878 English pun sentences, we selected 500 homographic and 500 heterographic puns with concrete concepts that are more easily visualized through images.

\bfsection{Generating Pun Explanation Images}
\modelname is designed to assess a VLM's capability to resolve lexical ambiguity with visual context.
In terms of a pun, the context should depict both meanings within the pun. Such images are hard to find among natural images due to their complex and sometimes ambivalent meanings. Further, such visual designs are typically proprietary, which contradicts our goal of an open-source dataset. Hence, we resort to creating new images that fit our requirements.

We recruited three NLP researchers to actively prompt the text-to-image generation model DALL-E 3~\cite{openai2023dalle3} to create images fitting our pun criteria while maintaining a natural appearance.
The base text-only dataset provided the puns as data seeds~\cite{miller2017semeval}.
While we allow relative freedom in the choice of prompts, the workers reported that DALL-E 3 typically produced satisfactory images with straightforward instructions, as illustrated in Figure~\ref{fig:image_gen_process}.
Thanks to DALL-E 3's multi-turn interface, the researchers could request further image revisions if the initial output was unsuitable.
On average, $\sim 24\%$ samples needed such multi-step modification. We obtained $1000$ \textit{pun explanation} images after this process.




\bfsection{Retrieving Pun Disambiguator Images}
\modelname offers an alternative visual context: per each pun, we attach two images that describe each meaning of the pun. These images disambiguate the pun and are intended to be used in the binary classification task of pun disambiguation explained in \cref{sec:task_definition}.

As a \textit{pun disambiguator} image is aligned to a single meaning, searching for the required image is easier compared to the \textit{pun explanation} images that require encoding both meanings in the same image. Hence, we opt for image retrieval from the LAION 2B web image-text dataset~\cite{schuhmann2022laion} rather than image generation. Using the CLIP~\cite{radford2021learning}-based image search API~\cite{beaumont-2022-clip-retrieval}, we retrieve ten images per the meaning of a pun. Then, we manually select the top one that best fits the description. We discard the whole sample when there is no suitable image. We considered two criteria when selecting the images: first, images that explicitly contain the meaning or the pun word itself as printed text are discouraged as such images reward OCR capability rather than general visual understanding. Second, images with watermarks are filtered out to avoid confusion.

\begin{table}[t]
    \centering
    \begin{tabular}{lcccc}
    \toprule
     & \multicolumn{2}{c}{\bf En$\rightarrow$Fr}
     & \multicolumn{2}{c}{\bf En$\rightarrow$De}
     \\
     & \multicolumn{2}{c}{Meaning}
     & \multicolumn{2}{c}{Meaning}
     \\
     Model & Freq$\uparrow$ & Freq$\downarrow$
     & Freq$\uparrow$ & Freq$\downarrow$
     \\
    \midrule
    GPT4 & 68.5 & 75.9 & 71.2 & 73.4 \\
    \xspace\xspace
    \xspace\xspace\xspace
    + Caption & 73.4 & 77.8 & 74.6 & 76.6 \\
    \bottomrule
    \end{tabular}
    \begin{tabular}{lcccc}
    \toprule
    & Accuracy (\%) & Cohen Kappa ($\kappa$) \\
    \midrule
    GPT4Eval & 78.1&  0.39 \\
    \bottomrule
    \end{tabular}
    \caption{Experiment on the effect of meaning frequency in puns. Top: division of pun reconstruction task results according to the commonality of meanings.
    Bottem: assessment of GPT-4-based meaning frequency ordering against an independent dataset with human-annotated meaning frequencies~\cite{rice2019comparison}.}
    \label{tab:meaning_freq}
\end{table}

\subsection{Translating Puns to Multilingual Targets}
\label{subsec:data_collection_translation}

Evaluating a machine's ability to understand puns is a complex task. Without a rule-based algorithm to measure this capability, the assessment often relies on human judgment or other machines. However, relying on human evaluation can limit the scalability of the assessment process, while machine-based evaluation, such as using models like GPT-4~\cite{openai2023gpt4}, may introduce undesirable biases~\cite{liu2023gpteval,hada2023large}.
To overcome these challenges, we suggest an alternative evaluation method via a downstream task in translation, intentionally aligning with previous research in the field of multimodal machine translation.

\bfsection{Translation with Machine Assistance}
We translate the original English pun sentence into three languages (German, French, and Korean).
Note that we should ensure that the ambiguity in English does not carry over into the translated targets.

We here design a cooperative framework between machines and humans for pun translation. 
Per each language pair (\eg En $\rightarrow$ De), we recruit a bilingual worker whose native language is the target language (\eg De). Frst, we use off-the-shelf translation models to generate three candidates.
Then, the human workers select the best one and make further modifications to finalize the translation.
This machine-assisted translation aligns with common practices in the industry~\cite{federico-etal-2012-measuring}.
We chose machine-human cooperation for two reasons: firstly, we saw that our human translators find pun translation difficult. Machine suggestions can serve as starting points here. Secondly, this method expedited the annotation process and reduced costs.

\bfsection{Addressing Lingering Ambiguity}
Certain cases arise where the ambiguity in the source language is retained in the translated text in literal translation. For example, consider the sentence: “A baseball player was a thief. He was always trying to steal.” The pun in this sentence relies on the dual meanings of “steal”—“to take without permission” and “to steal a base in baseball.” The challenge in translation is twofold:
Some languages contain equivalent idiomatic expressions (e.g., “stehlen” in German), which can result in similar ambiguities in the target text. To address this, translators were instructed to select alternative words that avoid unintended double meanings whenever possible.
The pun's humor is implied contextually within the first sentence, even if the pun word itself is not explicitly mentioned.
For such instances, indirect translations were permitted, allowing human translators to render distinct interpretations of the pun without preserving its exact wording. To further refine the outputs, we applied text-based deduplication to eliminate closely matching translations.
Refer to \cref{sec:appendix_data_collection} for more details. 

\subsection{Dataset Analysis}
\label{subsec:dataset_analysis}

Our pipeline yields a dataset comprising 500 homographic and 500 heterographic pun sentences, each accompanied by one \textit{pun explanation} image, two \textit{pun disambiguator} images, and translations to three languages.

\bfsectionnodot{How natural are the generated images?}
Given the limited availability of real-world images accurately depicting puns, we opted to use AI-generated visuals. To gauge the difference between generated and authentic images, we conducted two human evaluation studies, comparing our generated images against natural image-pun pairs sourced from the web (\url{https://www.reddit.com/r/puns/}).

In the first study, human evaluators were asked to identify the correct text pun associated with each image from a set of potential matches. Results showed that natural images achieved an accuracy of 86\%, while our generated images achieved a slightly higher accuracy of 92\%. This test was conducted using a set of 50 randomly selected images.
In the second study, we conducted an A/B comparison to assess the perceived naturalness of the images. To ensure consistency, natural images containing multiple panels, written text, or well-known characters were excluded from the evaluation. Across three independent evaluators, the naturalness test resulted in accuracy rates of 66\%, 72\%, and 74\%, respectively, using another set of 50 random images.
Overall, despite slight distributional differences between the generated and natural pun images, the disparity is considered acceptable. These findings indicate that evaluations performed within our benchmark can be reasonably extrapolated to real-world settings.

\bfsection{Common vs. uncommon meanings}
In \modelname, each sample contains a pun phrase with two distinct meanings. This section explores how the popularity, or frequency, of each meaning influences downstream performance.
To investigate, we rank the meanings of each word by their frequency using zero-shot GPT-4.
To ensure the accuracy of GPT-4's assessments, we cross-reference these with human-annotated frequency data from Rice et al.~\cite{rice2019comparison}, which includes 890 homonyms with annotated frequencies. The lower section of Table~\ref{tab:meaning_freq} compares GPT-4's frequency rankings with the human-annotated ground truth.
Next, using GPT-4, we categorize our data into two groups based on more and less frequent meanings.
This categorization is then analyzed through the \textit{pun reconstruction} task outlined in \cref{sec:task_definition}.
As illustrated in the upper part of Table~\ref{tab:meaning_freq}, the \textit{pun reconstruction} task reveals that inputs with common meanings present more challenges than those with uncommon ones when using GPT-4. This suggests that texts with an uncommon meaning supplement the model's inherent understanding of the more frequent meaning.
\begin{table}[t]
\centering
\resizebox{\columnwidth}{!}{%
\begin{tabular}{*{4}{c}}\toprule
Metric &Translation &Homo &Hetero \\\midrule
\multirow{2}{*}{Win Rate (\%)} &Plain &90.7 &82.1 \\
&Pun-aware &9.3 &17.9 \\
\midrule
\multirow{2}{*}{Score (Average)} &Plain & 94 & 93.9 \\
&Pun-aware & 88.8 & 87.8 \\
\bottomrule
\end{tabular}
}
\caption{Statistical differences between unconditional translation and pun-aware translation, averaged across languages. Text similarity was evaluated using BERTScore~\cite{zhang2019bertscore}.}
\label{tab:translation_compare}
\end{table}

\bfsectionnodot{How different are disambiguated translations from unconditional ones?}
When disambiguation is enforced as a strict criterion, the resulting translations are expected to differ from straightforward, unconditional translations. To quantify the extent of this difference, we compare the unconditional translation $\hat{y}_{0}$ against two baselines: (1) another unconditional translation produced by a different annotator ($\hat{y}_{1}$), and (2) the disambiguated translation ($y$). We measure text similarity scores for each pair: $s_1 = sim(\hat{y}_{0}, \hat{y}_{1})$ and $s_2 = sim(\hat{y}_{0}, y)$, and compute the win rate as the proportion of cases where $s_2$ exceeds $s_1$. The results, summarized in Table~\ref{tab:translation_compare}, show that although disambiguation instructions lead to noticeable changes, the overall difference remains relatively small. Further details can be found in ~\cref{sec:appendix_data_collection}.

\begin{table}[t]
    \centering
    \small
    \begin{tabular}{l@{\hskip 2mm}lcccc}
    \toprule
    & Model & Inputs & Homo & Hetero \\
    \midrule
    \parbox[t]{3mm}{\multirow{4}{*}{\rotatebox[origin=c]{90}{LM}}}
    & Vicuna & L &69.4 &	71.2 \\ 
    & Qwen-VL & L & 43.8& 57.8\\
    & LLaVA & L & 76.0 & 71.8\\
    & GPT-4 & L & 95.4 & 92.0 \\
    \midrule
    \parbox[t]{3mm}{\multirow{2}{*}{\rotatebox[origin=c]{90}{SM}}}
    & Vicuna & V + L & 74.6 ($\uparrow$ 5.2) & 76.6 ($\uparrow$ 5.4) \\
    & GPT-4 & V + L & 96.0 ($\uparrow$ 0.6) & 92.4 ($\uparrow$ 0.4) \\
    \midrule
    \parbox[t]{3mm}{\multirow{3}{*}{\rotatebox[origin=c]{90}{VLM}}}
    & Qwen-VL & V + L & 63.6 ($\uparrow$ \textbf{19.8})&70.8 ($\uparrow$ \textbf{13.0})\\
    & LLaVA & V + L & 81.8 ($\uparrow$ 5.8)&73.0 ($\uparrow$ 1.2)\\
    & GPT-4 & V + L & 97.6 ($\uparrow$ 2.2) & 94.0 ($\uparrow$ 2.0) \\
    \bottomrule
    \end{tabular}
    \caption{Results on the pun grounding task. We report the exact match accuracy of the generated pun phrase. $\uparrow$ denotes the performance gain from visual context.}
    \label{tab:pun_grounding}
\end{table}

\section{Task Overview}
\label{sec:task_definition}

We pose three multimodal pun understanding tasks on the collected annotations to test models' capability to use visual context in addressing lexical ambiguity, as illustrated in Figure~\ref{fig:main}.
Each task evaluates different aspects:
the easier \textit{Pun Grounding} task can be solved without image input.
It is aimed at determining if less advanced models, which might not fully resolve such challenges, can enhance their performance with added visual information.
The second task of \textit{pun disambiguation} is designed to necessitate the usage of visual context.
Finally, the \textit{pun reconstruction} task replicates a practical multimodal literacy scenario. 
This task necessitates that models not only use the given translation but also infer or extract the underlying pun meaning that the translation does not explicitly convey, potentially drawing on visual inputs to do so.

\bfsection{Pun Grounding}
The first step in understanding a pun is to identify it. Our initial task examines whether visual context aids models in identifying pun phrases within sentences.
Given the whole English sentence $x^i = [x^i_0, \ldots, x^i_t]$ containing a pun phrase $s^i = [x^i_k, \ldots, x^i_l]$ and its corresponding \textit{pun explanation} image $v^i_{e}$, the model returns a pun phrase candidate $\bar{s}^i$.
Note that while the actual target phrase $s^i$ is part of the full sentence $x^i$, the model's output $\bar{s}^i$ is not bound by this constraint. 
We purposefully formulate this task as a sequence-to-sequence problem to facilitate zero-shot evaluation across various baselines.
The model's output is then assessed for exact text match with the actual pun phrase to determine accuracy.

\bfsection{Pun Disambiguation}
Once models pinpoint a pun's location, they must then interpret its semantics. Understanding a pun hinges on recognizing the different meanings of the pun phrase, as its humor lies in this ambiguity. In this task, we assess the models' proficiency in correlating each meaning of the pun with its associated visual context.
Given the English sentence $x^i$ and the \textit{pun disambiguator} image $v^i_{d}$ aligned with one of the meanings constructing the pun,
the model should produce a translation of the sentence into a target language (\eg German $\bar{y}^i_{De}$).
Notably, the translated text should be free of any ambiguity stemming from the pun, closely aligning with the meaning depicted in the provided image.
We compare the model-generated translation $\bar{y}^{i,j}_{De}$ with two translation targets $y^{i,0}_{De},y^{i,1}_{De}$, each corresponding to a different meaning of the pun.
The model's output is considered correct if it more closely resembles the ground-truth translation $y^{i,j}_{De}$ that corresponds to the meaning depicted in the image $v^{i,j}_{d}, j \in {0, 1}$. Refer to \cref{subsec:exp_ours_results_pd} for the implementation details.


\bfsection{Pun Reconstruction}
The final task is to reconstruct the complete pun sentence.
To make the problem deterministic, we provide two types of inputs to the model:
a non-English language translation of the original pun sentence that has been clarified of any ambiguities (\eg German $y^{i,j}_{De}$) and the related \textit{pun explanation} image $v^i_{e}$.
The model then generates an output $\bar{x}^i$, which we compare with the original English pun sentence $x^i$ to determine if both English sentences encapsulate the same pun.
It is a complex task to determine whether two sentences contain the same pun, and we resort to machine-based evaluation with GPT-4 to obtain the binary decision.
We verify GPTEval's validity here using human evaluation in~\cref{sec:appendix_gpteval_test}.

\section{Experiments on \modelname benchmark}
\label{sec:ours_exp}

\begin{table*}[ht]
    \centering
    \begin{tabular}{l@{\hskip 1.3mm}l@{\hskip 1.3mm}c@{\hskip 1.3mm}ccccccccc}
    \toprule
    &  &  
    & \multicolumn{3}{c}{\bf En $\rightarrow$ De}
    & \multicolumn{3}{c}{\bf En $\rightarrow$ Fr}
    & \multicolumn{3}{c}{\bf En $\rightarrow$ Ko} \\
    & Model & Inputs
    & Homo & Hetero & All
    & Homo & Hetero & All
    & Homo & Hetero & All \\
    \midrule
    & Random & & 50.0 & 50.0 & 50.0 & 50.0 & 50.0 & 50.0 & 50.0 & 50.0 & 50.0 \\
    \midrule
    \parbox[t]{3mm}{\multirow{2}{*}{\rotatebox[origin=c]{90}{SM}}}
    & Vicuna & V + L & 59.4& 64.4& 61.9 &61.4 & \underline{72.2}& 66.8& 55.4& 55.2& 55.3\\
    & GPT-4 & V + L & \textbf{68.2} & \textbf{74.6} & \textbf{71.4} & \textbf{69.0} & 
    \textbf{76.8} & \textbf{72.9} & \textbf{65.4} & \textbf{66.2} & \textbf{65.8} \\
    \midrule
    \parbox[t]{3mm}{\multirow{3}{*}{\rotatebox[origin=c]{90}{VLM}}}
    & Qwen-VL & V + L & 60.7 & 64.4 & 62.6 & 61.7 & 71.4 & 66.5 & 55.4 & \underline{57.2}& 56.3 \\
    & LLaVA & V + L & \underline{65.1} & \underline{70.8} & \underline{68.0} & 61.1 & 70.6 & 65.8 & \underline{58.1} & 56.9 & \underline{57.5} \\
    & LLaVA-MMT & V + L & 63.5 & 68.0 & 65.7 & \underline{64.1} & 70.0 & \underline{67.0} & 56.6 & 56.1 & 56.4 \\
    \bottomrule
    \end{tabular}
    \caption{Experimental results on the pun disambiguation task. All scores are reported in terms of binary classification accuracy. The best scores are \textbf{bolded} and the second-best ones are \underline{underlined}.}
    \label{tab:pun_disambiguation}
\end{table*}

\subsection{Models}
\label{subsec:exp_ours_models}

\bfsection{LM}
To measure the effectiveness of multimodal modeling, we establish baselines using unimodal text-only language models.
We incorporate an open-source model (Vicuna-13B~\cite{vicuna2023}) and the advanced proprietary language model (GPT-4~\cite{openai2023gpt4}).
Furthermore, we appropriate a visual-language model, LLaVA, for a text-only scenario by inputting only text prompts without the images.
This approach assesses the concept of \textit{multimodal alignment tax}~\cite{chen2023polite} in the context of pun interpretation, implying that fine-tuning a model on visual data might impair its original linguistic capabilities.
We do not test LM baselines against \textit{pun disambiguation} as the task necessitates visual context.

\bfsection{SM (Socratic Models)}
SM~\cite{zeng2022socratic}, also called pipelining~\cite{bitton2023breaking}, is a two-staged framework extending text-only LMs to multimodal tasks by first encoding the multimodal context to textual descriptions.
To implement SMs, we employ the same language models as previously mentioned and use BLIP-2 OPT 2.7B~\cite{li2023blip} as the visual description generator to encode the images into textual captions.

\bfsection{VLM}
Monolithic visual-language models directly take the raw images and user queries as inputs to produce textual responses.
We employ two popular and high-performing VLMs for this purpose: LLaVA 1.5 13B~\cite{liu2023improved} and Qwen-VL-Chat 7B~\cite{bai2023qwen}. (We refer to Qwen-VL-Chat as Qwen-VL in result tables due to space constraints.)
For the tasks of \textit{pun disambiguation} and \textit{pun reconstruction}, we also introduce a machine translation baseline. We thus fine-tune LLaVA with the Multi30k multimodal machine translation dataset~\cite{elliott2016multi30k}, yielding the LLaVA-MMT variant. We choose LoRA~\cite{hu2021lora} over full fine-tuning for efficient implementation.

\subsection{Do Images Help Pun Grounding?}
\label{subsec:exp_ours_results_pg}

\bfsection{Metrics}
We report accuracy based on the equality of the model-estimated pun phrase and the ground-truth pun phrase.
To check the equality, we use the exact match of the surface text form and report the accuracy of the outputs. 

\bfsection{Results}
As anticipated, the incorporation of visual context led to a consistent improvement in pun grounding performance across all models, including Socratic Models and Visual-Language Models (refer to Table~\ref{tab:pun_grounding}).
Also, GPT-4, a stronger model, could solve the task even without visual context, verifying our original intention of proposing this task to test the helpfulness of visuals where the task is straightforward but the models are less capable.
For evaluation fairness, we employed a standard prompt template across all models (details in \cref{sec:appendix_prompts}).
Note that while careful prompt engineering can further improve the scores, our findings focus on understanding the role of visual context in realistic scenarios rather than extracting the maximum potential from each model.

\subsection{Can VLMs Disambiguate with Images?}
\label{subsec:exp_ours_results_pd}

\begin{table*}[ht]
\resizebox{\linewidth}{!}{
    \centering\small
    \begin{tabular}{l@{\hskip 1.2mm}lcccccccccc} \toprule
    \multicolumn{12}{c}{\bf De$\rightarrow$En} \\\midrule 
     &   &   &  \multicolumn{3}{c}{Homo}  &  \multicolumn{3}{c}{Hetero}  &  \multicolumn{3}{c}{All} \\ 
     &  Model  &  Inputs  &  Correct (\%)  &  Bleu-4  &  METEOR  &  Correct (\%)  &  Bleu-4  &  METEOR  &  Correct (\%)  &  Bleu-4  &  METEOR \\
    \midrule
    \parbox[t]{3mm}{\multirow{4}{*}{\rotatebox[origin=c]{90}{LM}}}
     &  Vicuna  &  L  &   27.9 & 28.8 & 56.6 & 16.0 & 29.1 & 65.1 & 22.0 & 29.0 & 60.9\\
     &  GPT-4  &  L  &  43.1 & 30.1 & 66.1 & 45.2 & 30.7 & 70.9 & 44.2 & 30.4 & 68.5\\
     &  Qwen-VL  &  L   &  30.3 & 29.4 & 58.8 & 20.3 & 30.0 & 66.7 & 25.3 & 29.7 & 62.8\\
     &  LLaVA  &  L    &  31.7 & 27.7 & 57.9 & 19.0 & 29.9 & 65.6 & 25.4 &28.8 & 61.8\\
    \midrule
    \parbox[t]{3mm}{\multirow{2}{*}{\rotatebox[origin=c]{90}{SM}}}
     &  Vicuna  &  V + L   &  35.0($\uparrow$7.1) & 25.6 & 51.7 & 19.1($\uparrow$3.1) & 26.3 & 57.5 & 27.1($\uparrow$5.1) & 26.0 & 54.6\\
     
     &  GPT-4  &  V + L  &  62.9($\uparrow$19.8) & 298 & 655 & 45.9($\uparrow$0.7) & 307 & 685 & 54.4($\uparrow$10.2) & 30.3 & 67.0\\
    \midrule
    \parbox[t]{3mm}{\multirow{4}{*}{\rotatebox[origin=c]{90}{VLM}}}
     &  Qwen-VL  &  V + L   &  34.3($\uparrow$4.0) & 28.5 & 54.2 & 19.9($\downarrow$0.4) & 29.7 & 58.2 & 27.1($\uparrow$1.8) & 29.1 & 56.2\\
     &  LLaVA  &  V + L  &    33.2($\uparrow$1.5) & 28.7 & 55.1 & 20.1($\uparrow$1.1) & 29.2 & 61.2 & 26.7($\uparrow$1.3) & 26.0 & 58.2 \\
     &  GPT-4  &  V + L   &  \textbf{65.2}($\uparrow$\textbf{22.1}) & 29.9 & 63.8 & 50.6($\uparrow$\textbf{5.4}) & 29.3 & 65.3 & 57.9($\uparrow$\textbf{13.7}) & 29.6 & 64.6\\
    \cmidrule{2-12}
     &  LLaVA-MMT  &  V + L    &  27.0  &  12.3  &  38.1 &  31.5  &  25.6  &  45.7 &  29.3  &  18.5  &  41.9 \\
     
    \midrule
    \midrule
    \multicolumn{12}{c}{\bf Fr$\rightarrow$En} \\\midrule 
    \parbox[t]{3mm}{\multirow{4}{*}{\rotatebox[origin=c]{90}{LM}}}
     &  Vicuna  &  L    &  28.7 & 28.0 & 57.6 & 19.2 & 29.5 & 66.6 & 24.0 & 28.8 & 62.1\\
     &  GPT-4  &  L   &  60.0 & 30.0 & 66.2 & 44.5 & 30.1 & 70.3 & 52.3 & 30.1 & 68.3\\
     &  Qwen-VL  &  L   &  31.5 & 29.2 & 59.2 & 19.9 & 30.4 & 67.7 & 25.7 & 29.8 & 63.5\\
     &  LLaVA  &  L   &  32.6 & 27.9 & 58.4 & 21.0 & 29.3 & 67.7 & 26.8 & 28.6 & 63.1\\
    \midrule
    \parbox[t]{3mm}{\multirow{2}{*}{\rotatebox[origin=c]{90}{SM}}}
     &  Vicuna  &  V + L   &  38.4($\uparrow$\textbf{9.7}) & 24.0 & 50.5 & 18.1($\downarrow$1.1) & 25.3 & 55.9 & 28.3($\uparrow$\textbf{4.3}) & 24.7 & 53.2\\
     &  GPT-4  &  V + L  &  63.6($\uparrow$3.6) & 29.2 & 65.1 & 45.2($\uparrow$0.7) & 30.7 & 68.1 & 54.4($\uparrow$2.1) & 30.0 & 66.6\\
    \midrule
    \parbox[t]{3mm}{\multirow{4}{*}{\rotatebox[origin=c]{90}{VLM}}}
     &  Qwen-VL  &  V + L   &  37.0($\uparrow$5.5) & 28.1 & 55.7 & 22.4($\uparrow$2.5) & 29.6 & 61.7 & 29.7($\uparrow$4.0) & 28.9 & 58.7\\
     &  LLaVA  &  V + L   &  34.3($\uparrow$1.7) & 28.5 & 55.3 & 23.7($\uparrow$\textbf{2.7}) & 29.6 & 63.3 & 29($\uparrow$2.2) & 29.1 & 59.3\\
     &  GPT-4  &  V + L   &  65.6($\uparrow$5.6) & 29.8 & 63.0 & 46.1($\uparrow$1.6) & 29.3 & 65.6 & 55.9($\uparrow$3.6) & 29.6 & 64.3 \\
    \cmidrule{2-12}
     &  LLaVA-MMT  &  V + L   &  33.3  &  12.9  &  39.3 &  27.0  &  24.3  &  43.2  &  30.2  &  17.8  &  41.3  \\
     
    \midrule
    \midrule
    \multicolumn{12}{c}{\bf Ko$\rightarrow$En} \\\midrule 
    \parbox[t]{3mm}{\multirow{4}{*}{\rotatebox[origin=c]{90}{LM}}}
     &  Vicuna  &  L    &  26.3 & 25.4 & 48.3 & 11.1 & 25.8 & 48.6 & 18.7 & 25.6 & 48.5\\
     &  GPT-4  &  L   &  62.7 & 30.9 & 69.5 & 41.8 & 29.5 & 65.5 & 52.3 & 30.2 & 67.5\\
     &  Qwen-VL  &  L   &  26.6 & 28.8 & 51.0 & 12.5 & 28.1 & 51.7 & 19.6 & 28.5 & 51.4\\
     &  LLaVA  &  L   &  27.9 & 25.4 & 55.0 & 11.9 & 25.5 & 50.6 & 19.9 & 25.5 & 52.8\\
    \midrule
    \parbox[t]{3mm}{\multirow{2}{*}{\rotatebox[origin=c]{90}{SM}}}
     &  Vicuna  &  V + L  &  31.9($\uparrow$5.6) & 20.3 & 38.7 & 16.6($\uparrow$5.5) & 20.3 & 35.4 & 24.3($\uparrow$5.6) & 20.3 & 37.1\\
     &  GPT-4  &  V + L  &  68.1($\uparrow$5.4) & 30.7 & 69.9 & 46.4($\uparrow$4.6) & 29.3 & 64.4 & 57.3($\uparrow$5.0) & 30.0 & 67.2\\
    \midrule
    \parbox[t]{3mm}{\multirow{4}{*}{\rotatebox[origin=c]{90}{VLM}}}
     &  Qwen-VL  &  V + L & 35.5($\uparrow$\textbf{8.9}) & 26.8 & 46.2 & 18.3($\uparrow$5.8) & 26.7 & 45.9 & 26.9($\uparrow$7.3) & 26.8 & 46.1\\
     &  LLaVA  &  V + L  &  30.2($\uparrow$2.3) & 23.4 & 41.3 & 16.4($\uparrow$5.0) & 23.1 & 41.0 & 23.3($\uparrow$3.4) & 23.3 & 41.2\\
     &  GPT-4  &  V + L  &70.2($\uparrow$7.5) & 30.1 & 65.7 & 52.3($\uparrow$\textbf{10.5}) & 29.5 & 61.3 & 61.3($\uparrow$\textbf{9.0}) & 29.8 & 63.5 \\
    \cmidrule{2-12}
    &  LLaVA-MMT  &  V + L  &  28.0  &  6.3  &  38.5  &   18.3  &  15.0  &  46.7  &  23.3  &  10.7  &  42.6  \\

    \bottomrule
    \end{tabular}
    }
    \caption{Outcomes for the pun reconstruction task, where $\uparrow$ and $\downarrow$ signify the performance change attributed to the inclusion of visual context.
    The model with the largest performance increase is marked \textbf{bold} in each language.
    }
    \label{tab:pun_reconstruction}
\end{table*}

\bfsection{Metrics}
We conduct a generative evaluation for the pun disambiguation test.
The task for the machines is to translate a given pun sentence into a target language, using the accompanying image as a guide to disambiguate the meaning of the pun phrase.
In this generative test, the model generates a sequence of text, which is then evaluated against two potential translation targets. 
The model's output is considered accurate if it aligns more closely with the translation that corresponds to the context of the provided image.
We use BERTScore~\cite{zhang2019bertscore} to measure the text similarity following the human evaluation results in~\cref{sec:appendix_bertscore}.

\bfsection{Results}
All the considered baselines have demonstrated their ability to disambiguate translation outputs based on visual context, as illustrated in Table~\ref{tab:pun_disambiguation}.
Both strengthening the language model (Vicuna \vs GPT-4) and improving visual context processing (Vicuna with image captions from BLIP-2 \vs LLaVA) led to more accurate disambiguation.
Still, comprehending puns in the textual form was a more decisive factor for pun disambiguation than a stronger visual understanding, as GPT-4 with image captions outperforms all other models.
Interestingly, fine-tuning with the Multi30k multimodal machine translation dataset~\cite{elliott2016multi30k} harmed the accuracy of visual alignment. 
The fine-tuned model (LLaVA-MMT) underperforms the zero-shot LLaVA in nearly all aspects, except in the English-to-French translation of heterographic puns.
This finding echoes previous research~\cite{futeral-etal-2023-tackling}, which suggests that multimodal machine translation datasets cannot properly evaluate multimodal literacy capability.

\subsection{Do Images Help Pun Reconstruction?}
\label{subsec:exp_ours_results_pr}

\bfsection{Metrics}
The pun reconstruction task involves machines using both the human-translated text and the image context to recreate the original pun sentence.
Then, the reconstructed pun is compared with the original sentence for consistency in puns.
Still, determining whether two sentences share the same pun is a complex task.
To tackle this, we use a machine-based evaluation method with GPT-4~\cite{openai2023gpt4} to determine if the puns in both sentences are equivalent.
To ensure the validity of this approach, known as GPTEval, we further compare it with human annotations in~\cref{sec:appendix_gpteval_test}.
Additionally, we report on common text evaluation metrics, such as Bleu-4~\cite{papineni2002bleu} and METEOR~\cite{banerjee2005meteor}—metrics widely used in the machine translation domain.

\bfsection{Results}
The results in Table~\ref{tab:pun_reconstruction} affirm that visual context significantly enhances machines' ability to reconstruct puns and manage their inherent ambiguity. For all tested models, the inclusion of images consistently improved the accuracy of pun reconstruction. The only exception was the weakest model in both language processing and visual comprehension (SM based on Vicuna).
Notably, unlike the main metric of correctness, the automatic text evaluation scores (Bleu-4 and METEOR) did not reflect a clear trend.
Through manual inspection of the generated outputs, we saw that such scores were more aligned with changes in the surface form of the text, which did not necessarily correlate with the accurate identification of puns.
This resonates with previous reports stating that such text scores are not fully effective outside of their original domain of machine translation~\cite{liu2016not}.

Due to the differences in their forms, models found it more challenging to reconstruct heterographic puns than homographic ones.
Notably, incorporating visual context in these more complex scenarios led to significant improvements.
Furthermore, the benefit of visual context became even more evident when dealing with Korean inputs; a language typically considered more divergent from English than either German or French.
This reinforces the idea that machines depend more on visual cues when tackling complex linguistic tasks.
Finally, as in the pun disambiguation task, the fine-tuned LLaVA-MMT suffered from a decline in performance compared to the zero-shot LLaVA. This further supports the notion that visual understanding is necessary to handle \modelname.


\section{Related Work}
\label{sec:related_Work}

\bfsection{Multimodal Machine Translation}
By integrating backtranslation as a downstream task, \modelname contributes to the literature on Multimodal Machine Translation (MMT), 
a widely studied area that extends neural machine translation with additional visual contexts~\cite{specia2016shared,elliott2017findings,barrault2018findings}.
Previous research argues that visual information can help resolve ambiguities in the source text~\cite{li2022valhalla,hatami2022analysing}.
However, the primary dataset for MMT, Multi30K~\cite{elliott2016multi30k}, has limited examples of such ambiguities, leading to questions about the use of MMT for assessing multimodal literacy capacity~\cite{elliott2018adversarial,wu2021good,futeral-etal-2023-tackling}.
Another benchmark counteracts this phenomenon with manual annotation~\cite{futeral-etal-2023-tackling,bawden2018evaluating}.
Nevertheless, this dataset is relatively small (155 samples) due to the difficulty in pinpointing ambiguities within sentences. Additionally, the benchmark is limited to classification models.


\bfsection{Computational Pun Understanding}
After early research~\cite{ritchie2005computational} pointed out ambiguity as a key in pun generation, numerous studies have investigated automatic pun generation regarding heterographic puns, which slackens the surface form identity requirement for each meaning of the pun~\cite{he2019pun,yu2020homophonic,mittal2022ambipun}.
Other research explored homographic pun generation which is based on multiple meanings of a polysemous word~\cite{yu2018neural,luo2019pun,tian2022unified}.
Recently, Sun~\etal~\cite{sun2022context} extended the pun generation problem to consider contextual cues.
We extend this line of research with multimodal understanding.

\bfsection{Visual-Language Models}
The field has seen rapid growth since Flamingo~\cite{alayrac2022flamingo} illustrated the advantages of applying large language models to the visual domain.
BLIP-2~\cite{li2023blip}, utilizing the OPT language model~\cite{zhang2022opt}, made significant strides in image captioning. 
The introduction of a stronger language model~\cite{touvron2023llama} further enabled prompt-based control of the models.
MiniGPT-4~\cite{zhu2023minigpt} and LLaVA~\cite{liu2023visual} pioneered the field of visual instruction tuning.
InstructBLIP~\cite{dai2023instructblip}, an extension of BLIP-2, improved its capability to follow instructions more accurately.
Further developments in this domain include other models such as LLAMA-Adapter~\cite{zhang2023llama} and Qwen-VL~\cite{bai2023qwen}.
Our research puts visual language models (VLMs) to the test regarding their multimodal literacy capabilities. 
\section{Conclusion}
\label{sec:conclusion}
We introduced \modelnamefancy, a new benchmark for the multimodal literacy capability.
Based on \modelname, we craft three tests to measure how machines can utilize visual context to resolve inherent ambiguity in puns.
Our findings indicate that machines can indeed leverage visual information to enhance their understanding of text, as shown by their improved performance across all tasks.

However, achieving human proficiency in multimodal literacy is still a challenge. While our results are encouraging, there remains a considerable gap in machine capability to fully grasp and interpret the intricate relationship between text and visuals, particularly in more complex tasks like \textit{pun reconstruction}.
Therefore, we envision \modelname as not only a platform for testing but also as a starting point for the development of future multimodal models to actively navigate and integrate information from multiple modalities.

\section*{Limitations and Ethical Considerations}
\label{sec:limitations}

\modelname, while being a multilingual dataset, is built on the English-only pun corpus~\cite{miller2017semeval}.
As such, it primarily models lexical ambiguities unique to English, stemming from polysemies or similar surface forms of the language.
To enhance its linguistic diversity and applicability, expanding the dataset to include ambiguities inherent in other languages would be beneficial.
Such expansion would not only diversify the linguistic challenges in the dataset but also offer deeper insights into how lexical ambiguities manifest differently across various languages and cultures.

Although \modelname's size is much larger than that of the previous multimodal literacy dataset that features explicit ambiguities~\cite{futeral-etal-2023-tackling}, its total size is insufficient for creating a training split suitable for fine-tuning.
This limitation stems from the scarcity of puns, which are inherently challenging for humans to create as well and are not readily available in large quantities online. 
We thus plan to expand the dataset for multilingual puns in the future.

\bfsection{Ethical Considerations}
\modelname, constructed using existing English puns, may inadvertently perpetuate cultural biases and stereotypes present within the humor. Although human annotators were instructed to eliminate any puns expressing explicit hatred, subtle biases can still be perpetuated through seemingly innocuous humor.

To address ethical concerns in the data curation process, we confirmed that all human annotators either volunteered willingly or were compensated fairly for their contributions. 
We defer the details to~\cref{sec:appendix_data_collection}.

\section*{Acknowledgment}
\label{sec:ack}

This work was partly supported by an IITP grant funded by the Korean Government (MSIT) (No. RS-2020-II201361, Artificial Intelligence Graduate School Program (Yonsei University) and RS-2024-00353131) and the National Research Foundation of Korea (NRF) grant funded by the Korea government (MSIT) (No. RS-2024-00354218).

%
%
\bibliography{main}

\clearpage
\appendix
\setcounter{page}{1}

\section{Hyperparameters \& Setup}
\label{sec:appendix_hyperparams}

\bfsection{Models}
In all GPT-4~\cite{openai2023gpt4} usage, we use the \url{gpt-4-0613} endpoint.
When conducting experiments with open-source models, we leverage the official implementation codes in conjunction with publicly available weights from the Huggingface Hub (\url{https://huggingface.co/models}).
For our work with the Vicuna~\cite{vicuna2023} language model, we employed the \url{lmsys/vicuna-7b-v1.5} endpoint. Additionally, for the LLaVA 1.5 13B~\cite{liu2023improved} and Qwen-VL-Chat~\cite{bai2023qwen} visual-language models, we used the following model parameters: \url{liuhaotian/llava-v1.5-13b} and \url{Qwen/Qwen-VL-Chat}, respectively.

\bfsection{Text Generation}
We use deterministic greedy sampling for all experiments, introducing no randomness or external hyperparameter in the text generation process. Except for GPT-4, all models were allowed to generate up to 200 tokens with the freedom of an early stopping.

\bfsection{Computational Resources \& Fine-tuning}
We used the OpenAI API for inferring GPT-4~\cite{openai2023gpt4} outputs. For open-source models such as Vicuna~\cite{vicuna2023}, LLaVA~\cite{liu2023improved}, and Qwen-VL-Chat~\cite{bai2023qwen}, we use a single NVIDIA A100 40GB GPU for inference. While the exact inference speed varies depending on the length of prompts and responses, a query takes about $\sim0.8$ seconds to terminate when utilizing batch processing.
Fine-tuning LLaVA was also possible in a single A100 40GB GPU thanks to the efficient LoRA-based implementation~\cite{hu2021lora}. We trained each translation model for ten epochs in the training split of the Multi30k dataset~\cite{elliott2016multi30k} with early stopping, which took $\sim20$ hours on average.

\section{Data Collection Details}
\label{sec:appendix_data_collection}

\bfsection{Generating Pun Explanation Images}
We rely on the DALL-E 3~\cite{openai2023dalle3} image-to-text model to generate images that explain both meanings of the pun at the same time.
DALL-E 3 can be accessed either from the GPT-4 (\url{https://chat.openai.com/}) or Bing Image Generator (\url{https://www.bing.com/create}) web interface.
Our annotators employed both interfaces.

In contrast to previous studies that employed designers for image generation tasks involving machine-human collaboration~\cite{bitton2023breaking}, we chose to recruit three NLP researchers to generate images with DALL-E 3 and curate the outputs. This decision was driven by the necessity for an accurate representation of the puns' meanings rather than the artistic quality of the images, considering the specific requirements of our research task.
The NLP researchers participated voluntarily in this annotation task.

\bfsection{Translating Puns}
To streamline the translation process while ensuring clarity in meaning, we adopted a machine-assisted translation approach. This method simplifies the task for human annotators, who are required to ensure that the translations clearly reflect the intended meaning of the text. Initially, we provide three machine-generated translation options for the annotators to select and refine. These options are created using GPT-4, which we prompt in two distinct ways, and DeepL (\url{https://www.deepl.com/translator}), a proprietary translation service.

\begin{figure*}
    \centering
    \includegraphics[width=1.0\textwidth] {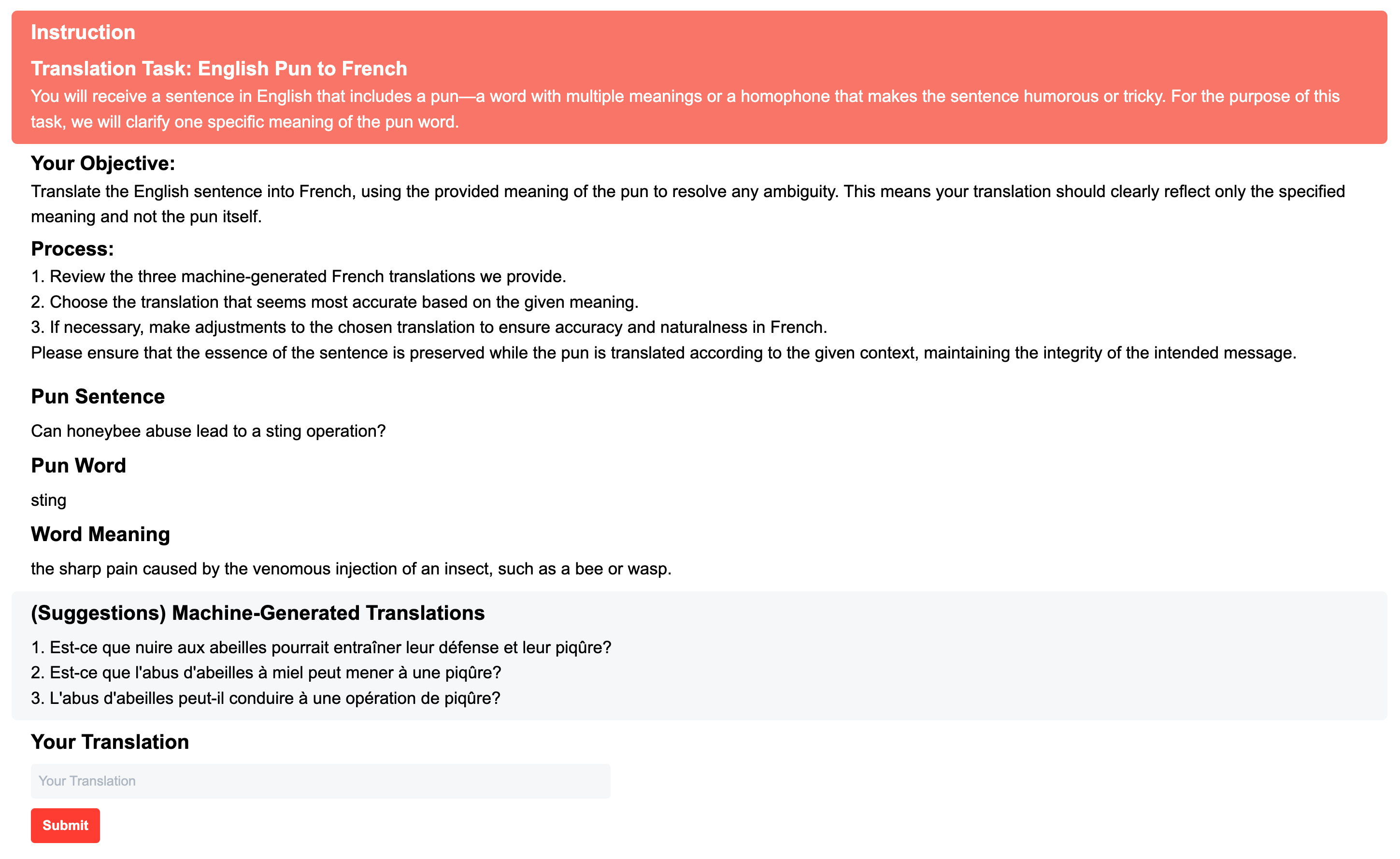}
    \caption{A screenshot of the human annotation interface for pun-aware text translation.}
    \label{fig:ax_translation_screen}
\end{figure*}

Human annotators then use these machine-provided translations as a base to craft the final version of the translated pun sentence, as depicted in the user interface shown in Figure~\ref{fig:ax_translation_screen}. For each of the three target languages (German, French, and Korean), we engaged a native speaker who is also proficient in English, ensuring both linguistic accuracy and fidelity to the original pun's meaning. The extent for how much the translations written by human annotators were drifted from unconditional translation is reported in Table~\ref{tab:translation_compare_full}. We pay each annotator 12–15\$ per hour. We have received approval from the university department and conducted data collection.

\begin{table*}[ht]
\centering
\resizebox{0.9\textwidth}{!}{%
\begin{tabular}{*{8}{c}}\toprule
& &\multicolumn{2}{c}{\textbf{De}} &\multicolumn{2}{c}{\textbf{Fr}} &\multicolumn{2}{c}{\textbf{Ko}} \\\cmidrule{3-8}
Metrics &Translation &Homo &Hetero &Homo &Hetero &Homo &Hetero \\\midrule
\multirow{2}{*}{Win Rate (\%)} &Plain & 93.2 & 80.0 & 92.4 & 81.0 & 86.4 & 85.4 \\
&Pun-aware & 6.8 & 20.0 & 7.6 & 19.0 & 13.6 & 14.6 \\
\multirow{2}{*}{Score (Average)} &Plain & 95.2 & 95.0 & 95.5 & 95.4 & 91.3 & 91.4 \\
&Pun-aware & 84.5 & 88.0 & 96.6 & 89.6 & 85.0 & 85.9 \\
\bottomrule
\end{tabular}
}
\caption{Statistical differences between unconditional translation and pun-aware translation. Text similarity was evaluated using BERTScore~\cite{zhang2019bertscore}.}

\label{tab:translation_compare_full}
\end{table*}

\section{Testing the Validity of BERTScore}
\label{sec:appendix_bertscore}

\begin{table}[t]
    \centering
    \begin{tabular}{l|cc|cc}
    \toprule
    & \multicolumn{2}{c}{LLaVA}
    & \multicolumn{2}{c}{GPT4} \\
    Method & Acc (\%) & $\phi$  & Acc (\%) & $\phi$ \\
    \midrule
    BLEU & 84 & 57 & 84 & 59 \\
    METEOR & 82 & 52 & 80 & 49 \\
    Rouge-L & 84 & 41 & 82 & 54 \\
    BERTScore & \textbf{93} & \textbf{81} & \textbf{89} & \textbf{72} \\
    GPTEval & 76 & 41 & 76 & 43 \\
    \bottomrule
    \end{tabular}
    \caption{We compare each metric with human judgments on a set of 100 samples. $\phi$ denotes the Phi correlation coefficient.}
    \label{tab:ax_bertscore}
\end{table}

In the Pun Disambiguation task, we require the models to generate the disambiguated text. Thus, we need an automatic algorithm to decide whether the generated output aligns with the intended pun meaning. We formulate this as a text-only problem of matching the output with the ground-truth disambiguated translation result.

We consider various options here: three translation metrics (Bleu, METEOR, and Rouge-L) and two model-based metrics (BERTScore~\cite{zhang2019bertscore} and GPT-based Evaluation). Note that GPTEval here does not receive images as inputs following the other options.
Given the generation output of a VLM (LLaVA) or a Socratic Model (GPT4), we ask human annotators for a ternary classification: Match, No Match, and Invalid. A sample is classified as matched if it is better aligned to the intended translation target than the other one and as not matched vice versa. We also let the annotators mark invalid outputs in which the text does not align with either target. Finally, we compare the human decisions with automatic algorithms on 100 valid samples. We filtered out 32 invalid outputs from LLaVA and none from GPT4.

The results show that BERTScore greatly improves over the traditional translation metric baselines. As the disambiguation of a pun sentence typically lies in the correct translation of salient phrases, conventional metrics without semantic understanding are not sufficient for the task. On the other hand, BERTScore shows an acceptable correlation with human annotations. We thus employ it as the metric for the Pun Disambiguation task.
Perhaps surprisingly, the strong LLM backbone of GPTEval yields the worst outcome. Upon qualitative examination, we saw that GPT tends to favor certain targets regardless of the input. We leave alleviating this bias of GPTEval to future research.

\begin{table*}[ht]
    \centering
    \begin{tabular}{lcccccccccc}
    \toprule
    & \multicolumn{5}{c}{Multimodal}
    & \multicolumn{5}{c}{Text-Only} \\
    & Precision & Recall & F1 & Acc (\%) & $\rho$
    & Precision & Recall & F1 & Acc (\%) & $\rho$ \\
    \midrule
    De $\rightarrow$ En & 
    0.91 & 0.95 & 0.93 & 89 & 0.65 &
    0.91 & 0.92 & 0.92 & 87 & 0.62 \\
    Fr $\rightarrow$ En & 
    0.92 & 0.92 & 0.92 & 88 & 0.62 &
    0.91 & 0.90 & 0.91 & 86 & 0.62 \\
    Ko $\rightarrow$ En & 
    0.84 & 0.86 & 0.85 & 80 & 0.54 &
    0.88 & 0.88 & 0.88 & 85 & 0.66 \\
    \bottomrule
    \end{tabular}
    \caption{Test results of machine-based evaluation scheme using GPT-4 in the pun reconstruction task. We compare GPT-4's decision with human judgments on a set of 100 samples. $\rho$ denotes the Pearson correlation coefficient.}
    \label{tab:ax_gpteval_test}
\end{table*}

\section{Testing the Validity of GPTEval}
\label{sec:appendix_gpteval_test}

To ensure the reliability of our machine-based evaluation method, we conducted a comparative analysis with human assessments. This involved manually annotating 100 sample outputs from a multimodal approach (GPT-4 with image captions) and a text-only approach (GPT-4). 
The findings in Table~\ref{tab:ax_gpteval_test} indicate that GPT-4's evaluations can be considered dependable for assessing the performance of machine-generated outputs in the pun reconstruction task.
\section{Prompt Templates}

\label{sec:appendix_prompts}
\subsection{Pun Grounding}

\begin{itemize}
\item \textbf{LM} \\

\hspace{0.0em}
\begin{minipage}{\linewidth}

\small

\begin{lstlisting}
  ###Variables
   PUN_SENTENCE
  
  ###PROMPT
   [sentence]: {PUN_SENTENCE}
   This is a pun sentence. Identify the specific word or phrase 
   that creates the pun. Respond with only the word or phrase that 
   makes it a pun, without any explanation.
   [answer]:
\end{lstlisting}
\end{minipage}

\item \textbf{SM or VLM} \\

\hspace{0.0em}
\begin{minipage}{\linewidth}
\small
\begin{lstlisting}
  ###Variables
   PUN_SENTENCE, IMAGE
  
  ###PROMPT
   [sentence]: {PUN_SENTENCE}
   This is a pun sentence. Identify the specific word or phrase
   that creates the pun, given the image as context "{IMAGE}". 
   Respond with only the word or phrase that makes it a pun, 
   without any explanation.
   [answer]:
\end{lstlisting}
\end{minipage}
\end{itemize}


\subsection{Pun Disambiguation}
\begin{itemize}

\item \textbf{SM or VLM} \\

\hspace{0.0em}
\begin{minipage}{\linewidth}
\small
\begin{lstlisting}
  ###Variables
   PUN_SENTENCE, IMAGE, LANGUAGE
  
  ###PROMPT
   [English]: {PUN_SENTENCE}
   Translate the given English sentence into {LANGUAGE}, given the 
   image as context "{IMAGE}". Please respond using the format 
   below:
   [{LANGUAGE}]:
\end{lstlisting}
\end{minipage}
\end{itemize}

\subsection{Pun Reconstruction}

\begin{itemize}
\item \textbf{LM} \\

\hspace{0.0em}
\begin{minipage}{\linewidth}
\small

\begin{lstlisting}
  ###Variables
   PUN_SENTENCE_IN_LANGUAGE, LANGUAGE
  
  ###PROMPT
   Please translate the following sentence from {LANGUAGE} into 
   English, ensuring that the translation contains a pun. I will 
   provide you with a sentence in {LANGUAGE} using the format 
   [{LANGUAGE}]: "{PUN_SENTENCE_IN_LANGUAGE}". Please respond using 
   the format below:
   [English]:
\end{lstlisting}
\end{minipage}

\item \textbf{SM or VLM} \\

\hspace{0.0em}
\begin{minipage}{\linewidth}
\small

\begin{lstlisting}
  ###Variables
   PUN_SENTENCE_IN_LANGUAGE, LANGUAGE, IMAGE
  
  ###PROMPT
   Please translate the following sentence from {LANGUAGE} into
   English, ensuring that the translation contains a pun. I will
   provide you with a sentence in {LANGUAGE} using the format
   [{LANGUAGE}]: "{PUN_SENTENCE_IN_LANGUAGE}".
   
   [Image Description]: {IMAGE}
   This image description is about two meanings of the word that 
   you are expected to create. Use this information to craft your
   pun-inclusive English translation. Please respond using the
   format below:
   [English]:
\end{lstlisting}
\end{minipage}
\end{itemize}



\begin{figure*}[!ht]
    \centering
    \includegraphics[width=1.0\textwidth]{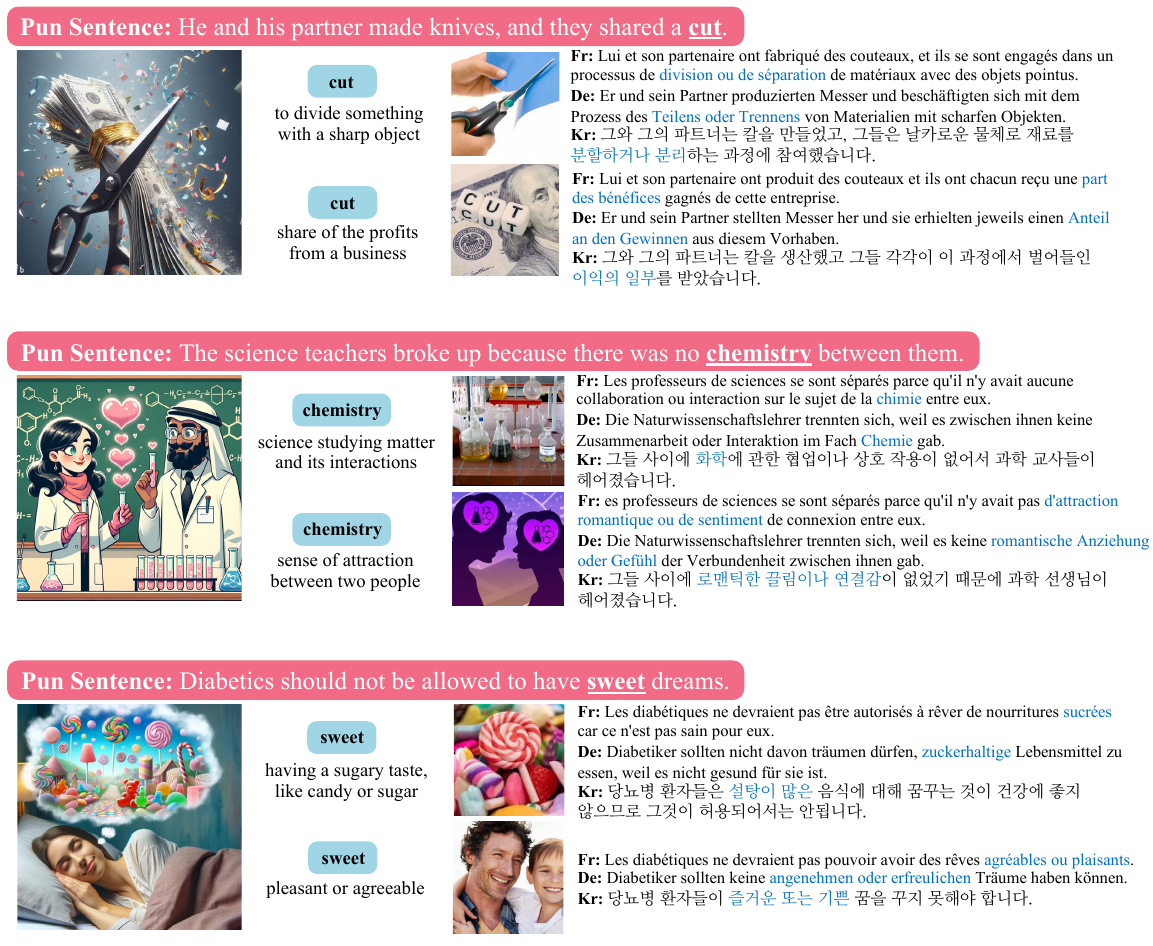}
    \caption{Example annotations of homographic puns in \modelnamefancy benchmark.}
    \label{fig:homo_samples}
\end{figure*}

\begin{figure*}[!ht]
    \centering
    \includegraphics[width=1.0\textwidth]{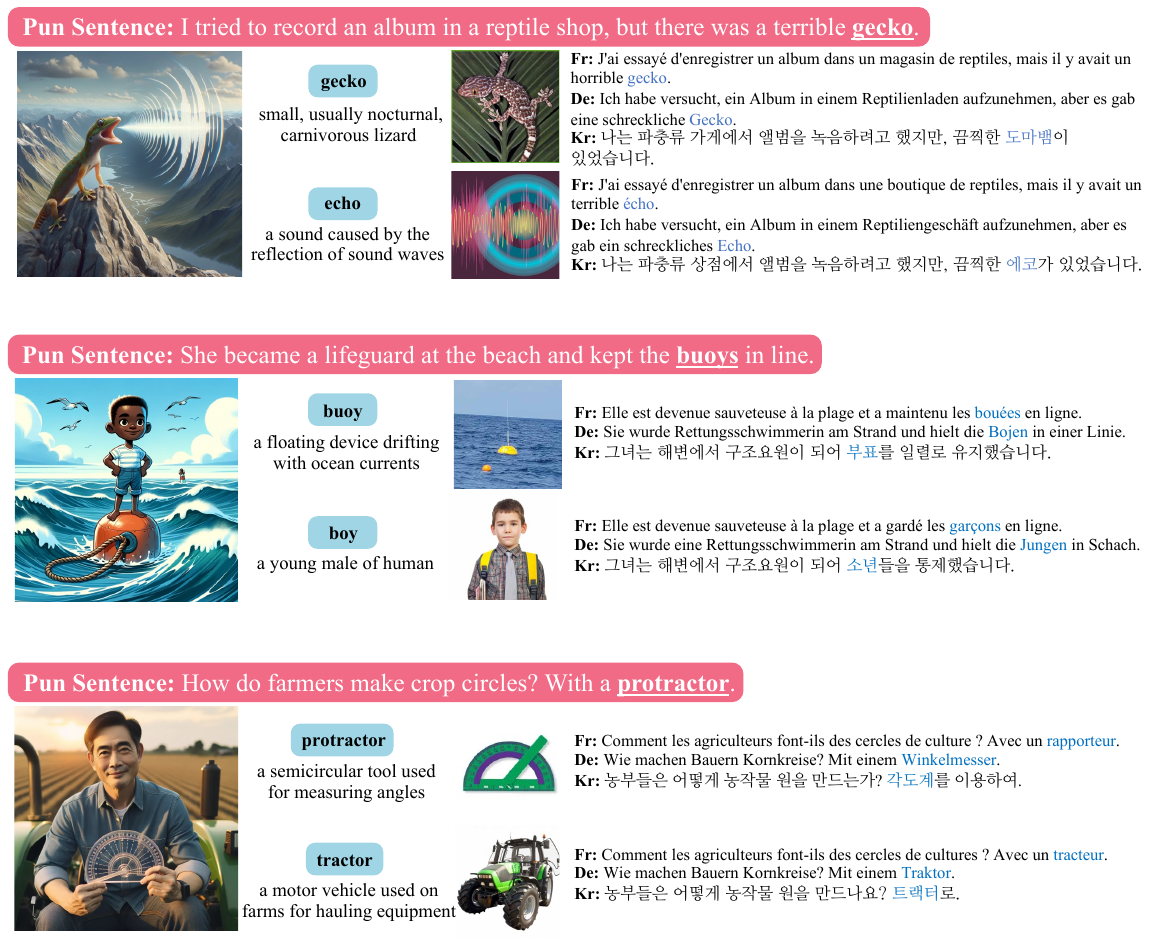}
    \caption{Example annotations of heterographic puns in \modelnamefancy benchmark.}
    \label{fig:hetero_samples}
\end{figure*}

\end{document}